\documentclass{article}

\PassOptionsToPackage{numbers, compress}{natbib}
\usepackage[final]{neurips_data_2021}

\usepackage[utf8]{inputenc} 
\usepackage[T1]{fontenc}    
\usepackage{hyperref}       
\usepackage{url}            
\usepackage{wrapfig}
\usepackage{booktabs}       
\usepackage{amsfonts}       
\usepackage{nicefrac}       
\usepackage{microtype}      
\usepackage{xcolor}         
\usepackage{graphicx}
\usepackage{bm}
\def\eg{\emph{e.g., }}

\def\ie{\emph{i.e., }}

\usepackage{multirow}
\usepackage{makecell}
\usepackage{multicol}

\usepackage{booktabs}
\usepackage{bbm}
\usepackage{multicol}
\usepackage{makecell}
\usepackage{xcolor}
\usepackage{xspace}
\usepackage{float}
\usepackage{color}
\usepackage{colortbl}
\usepackage{amsmath,amssymb} %

\renewcommand{\Re}{\mathbb{R}}
\newcommand{\hy}{\hat{y}}
\newcommand{\hb}{\hat{b}}
\newcommand{\ha}{\hat{a}}
\newcommand{\hp}{\hat{p}}

\renewcommand{\Sigma}{\mathfrak{S}}

\newcommand{\indic}[1]{\mathds{1}_{\{#1\}}}

\newcommand{\bloss}[1]{{\cal L}_{\rm box}(#1)}
\newcommand{\aloss}[1]{{\cal L}_{\rm angle}(#1)}

\newcommand{\hloss}[1]{{\cal L}_{\rm Hungarian}(#1)}

\newcommand{\lmatch}[1]{{\cal L}_{\rm match}(#1)}

\def\eqref#1{equation~\ref{#1}}

\def\1{\bm{1}}

\DeclareMathAlphabet{\mathsfit}{\encodingdefault}{\sfdefault}{m}{sl}
\SetMathAlphabet{\mathsfit}{bold}{\encodingdefault}{\sfdefault}{bx}{n}

\DeclareMathOperator*{\argmin}{arg\,min}

\newcommand{\noobject}{\varnothing}

\definecolor{mygray}{gray}{.92}

\newcommand{\green}[1]{\textcolor[RGB]{96,177,87}{#1}}

\renewcommand{\texttt}[1]{ $ {{\tt #1} } $}  

\usepackage[ruled]{algorithm2e}         
\usepackage{array}
\newcolumntype{I}{!{\vrule width 3pt}}
\newlength\savedwidth
\newcommand\whline{\noalign{\global\savedwidth\arrayrulewidth
                           \global\arrayrulewidth 2pt}%
                  \hline
                  \noalign{\global\arrayrulewidth\savedwidth}}
\newlength\savewidth
\newcommand\shline{\noalign{\global\savewidth\arrayrulewidth
                           \global\arrayrulewidth 0.5pt}%
                  \hline
                  \noalign{\global\arrayrulewidth\savewidth}}

\usepackage{dsfont}

\title{A Bilingual, Open World Video Text Dataset and End-to-end Video Text Spotter with Transformer}

\author{
  
  Weijia Wu\thanks{Equal contribution.} \,\thanks{This work was done when Weijia Wu were interns in MMU, Kuaishou Technology, Beijing, China.}\\
  Zhejiang University\\
  \texttt{weijiawu@zju.edu.cn} \\
  \And
  Yuanqiang Cai$^*$ \\
  Beijing University of Posts and Telecommunications \\
   \texttt{caiyuanqiang@bupt.edu.cn } \\
  \AND
  Debing Zhang \\
  Kuaishou Technology \\
  \And
  Sibo Wang \\
  Kuaishou Technology \\
  \And
  Zhuang Li \\
  Kuaishou Technology \\
  \And
  Jiahong Li \\
  Kuaishou Technology \\
  \And
  Yejun Tang \\
  Kuaishou Technology \\
  \And
  Hong Zhou\thanks{Corresponding author.} \\
  Zhejiang University \\
}

\begin{document}

\maketitle
\begin{abstract}
Most existing video text spotting benchmarks focus on evaluating a single language and scenario with limited data.
In this work, we introduce a large-scale, \textbf{B}ilingual, \textbf{O}pen World \textbf{V}ideo text benchmark dataset(BOVText). There are four features for BOVText. 
Firstly, we provide \textbf{2,000+} videos with more than \textbf{1,750,000+} frames, \textbf{25} times larger than the existing largest dataset with incidental text in videos. 
Secondly, our dataset covers \textbf{30+} open categories with a wide selection of various scenarios, \eg{\textit{Life Vlog, Driving, Movie, etc}}. 
Thirdly, abundant text types annotation~(\ie{} \textit{title, caption or scene text}) are provided for the different representational meanings in video. 
Fourthly, the BOVText provides bilingual text annotation to promote multiple cultures' live and communication. 

Besides, we propose an end-to-end video text spotting framework with Transformer, termed TransVTSpotter, which solves the multi-orient text spotting in video with a simple, but efficient attention-based query-key mechanism. 
It applies object features from the previous frame as a tracking query for the current frame and introduces a rotation angle prediction to fit the multi-orient text instance. On ICDAR2015(video), TransVTSpotter achieves the state-of-the-art performance with \textbf{44.1\%} MOTA, \textbf{9} fps.
The dataset and code of TransVTSpotter can be found at \href{https://github.com/weijiawu/BOVText-Benchmark}{\color{blue}{$\tt github.com/weijiawu/BOVText$}} and \href{https://github.com/weijiawu/TransVTSpotter}{\color{blue}{$\tt github.com/weijiawu/TransVTSpotter$}}, respectively.

\end{abstract}

\section{Introduction}

Text spotting~\cite{long2021scene,jung2004text} has received increasing attention due to its numerous applications in computer vision, \textit{e.g.,} document analysis, image-based translation, image retrieval~\cite{schroth2011exploiting,mishra2013image}, etc.
With the advent of deep learning and abundance in digital data, reading text from static images has made extraordinary progress in recent years with a lot of great public datasets~\cite{synthtext,karatzas2015icdar,totaltext} and algorithms~\cite{psenet,zhou2017east,lyu2018mask,liu2018fots}. 
By contrast, video text spotting almost remains at a standstill for the lack of large-scale multidimensional practical datasets, which limited numerous applications of video text, \eg{video understanding~\cite{srivastava2015unsupervised}, video retrieval~\cite{dong2021dual}, video text translation, and license plate recognition~\cite{anagnostopoulos2008license}, etc.}
%

\begin{figure}[t]
\begin{center}
\includegraphics[width=1\textwidth]{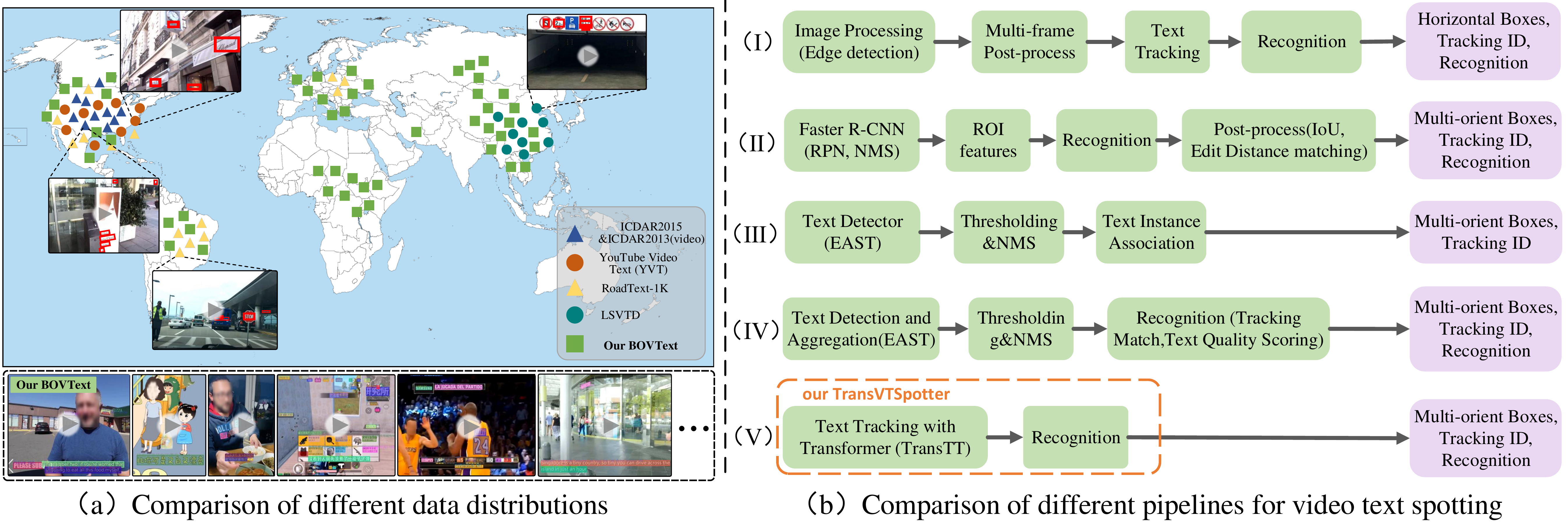}
\vspace{-7mm}
\caption{\textbf{Comparison of dataset distribution and pipeline}. (a) Data distribution. BOVText provides various open world scenarios with unique \textit{NBA, Game, etc.} (b) Pipelines: (\uppercase\expandafter{\romannumeral1}) Multi-stage pipeline in \cite{xi2001video}, \cite{hua2002efficient}, \cite{yi2009using} \textit{etc}; (\uppercase\expandafter{\romannumeral2}) Multi-orient video text spotting pipeline with Faster R-CNN~\cite{ren2015faster} proposed by Wang \textit{et al.}~\cite{wang2017end}; (\uppercase\expandafter{\romannumeral3}) Online text tracking pipeline in Yu \textit{et al.}~\cite{yu2021end}; (\uppercase\expandafter{\romannumeral4}) Fast video text spotting pipeline in Cheng \textit{et al.}~\cite{cheng2019you}; (\uppercase\expandafter{\romannumeral5}) An end-to-end pipeline with transformer in this work.} 
\label{comparison}
\end{center}
\vspace{-7mm}
\end{figure}

%
%
Video text spotting(VTS) is the task that requires simultaneously classifying, detecting, tracking and recognizing text instances in a video sequence.
There have been a few previous works~\cite{yang2017tracking,wu2015new} and datasets~\cite{nguyen2014video,karatzas2013icdar} in the community for attempting to develop video text spotting.
ICDAR2015~(Text in Videos)~\cite{karatzas2015icdar} was introduced during the ICDAR Robust Reading Competition in 2015 and mainly includes a training set of 25 videos~(13k frames) and a test set of 24 videos~(14k frames). The videos were categorized into seven scenarios: walking outdoors, searching for a shop in a shopping street, etc. 
YouTube Video Text~(YVT)~\cite{nguyen2014video} dataset harvested from YouTube, contains 30 videos with 13k frames, 15k for training, and 15k for testing. The text content in the dataset mainly includes overlay text and scene text (\eg{street signs, business signs, words on shirt}).
RoadText-1K~\cite{reddy2020roadtext} are sampled from BDD100K~\cite{yu2018bdd100k}, includes 700 videos~(210k frames) for training and 300 videos for testing. The texts are all obtained from driving videos and match for driver assistance and self-driving systems.
LSVTD~\cite{cheng2019you} includes 100 text videos, 13 indoor (\eg{ bookstore, shopping mall}) and 9 outdoor (\eg{highway, city road}) scenarios.
However, as shown in Figure.~\ref{comparison} (a), most existing video text datasets are limited by the amount of training data~(less than 300k frames), single video scenarios, and a single language. There are only a few outdoor scene text videos with 13k frames in ICDAR2015~(video). Similar situation for YVT, RoadText-1k, and LSVTD, the training set is limited and the dataset scenarios are single. This makes it difficult to evaluate the effectiveness of more advanced deep learning models for more open scenarios, such as \textit{game, sport and news report}.
Besides, most existing video text datasets are proposed before 2019 years, and some of them are no longer being maintained without an open-source evaluation script. 
The download links of YVT even have become invalid, which is not conducive to the development of video text spotting.
%

\vspace{-2mm}

In this work, we contribute a large-scale, bilingual open-world benchmark dataset~(BOVText) to the community for developing and testing video text spotting that can fare in a realistic setting. Our dataset has several advantages.
\textbf{Firstly}, the large training set~(\ie{2,000+video and 1,750,000+ video frames}) from \textit{KuaiShou} and \textit{YouTube} enables the development of deep design specific for video text spotting.
\textbf{Secondly}, unlike the existing datasets, BOVText support \textbf{30+} open scenarios, including many new scenarios such as \textit{Sportscast(NBA, FIFA World Cup...), Life Vlog, Game, etc}, as shown in Figure.~\ref{comparison}~(a). These data is collected from the worldwide user of \textit{YouTube}\footnote{https://www.youtube.com/} and \textit{KuaiShou}\footnote{https://www.kuaishou.com/en}, cover various daily scenarios without region limitation and virtual scenes. 
But the previous video text datasets usually are collected toward a special city or language from the hand-held camcorder.
\textbf{Thirdly}, BOVText is the first benchmark for supporting abundant text types annotation. 
Caption, title, and scene text are separately tagged for the different representational meanings in the video. This made our BOVText has the potential to promote other video-and-language tasks, such as video understanding.
\textbf{Fourthly,} bilingual text annotation(\ie{Chinese, English}) is provided in BOVText to promote multiple cultures' live and communication.

Except for the promising benchmark, we also proposed a simple, but effective video text spotter with transformer~(TransVTSpotter). 
As shown in Figure.~\ref{comparison}~(b), unlike previous methods that involve multiple steps, such as proposal generation, text aggregation, and post-processing(NMS),
TransVTSpotter only requires two steps. 1) Text tracking: for each consecutive frame image, we obtain the multi-orient boxes tracking trajectories of text by boxes IoU matching between the predicted detection boxes~\cite{carion2020end} and the predicted tracking boxes~\cite{sun2020transtrack}, where the detection boxes are obtained by taking an object query as input, just like DETR~\cite{carion2020end}. 
And features from previously detected objects to form another “track query” to discover associated objects~(\ie{} the predicted tracking boxes) on the current frames. Besides, an additional angle loss of multi-orient box and \textit{Hungarian angle cost} are design to obtain the angle of multi-orient. 2) Text recognition: recognizing the tracked texts with attention-based text recognizer~\cite{lu2021master}.
Without bells and whistles, TransVTSpotter achieves state-of-the-art performance on ICDAR2015 with \textbf{44.1\%} MOTA, \textbf{9} fps.
The main contributions of this work are three folds:

(1) We propose a large-scale, bilingual and open world video text spotting benchmark named BOVText. The proposed dataset provides \textbf{2,000+} videos, \textbf{1,750,000} frames, open videos scenarios~(\eg{}\textit{ Indoor, Outdoor, Game, Sport}), abundant text types~(\ie{} \textit{title, caption or scene text}), multi-stage tasks and is \textbf{25} times the existing largest dataset with incidental text.

(2) Caption, title, scene text, and other overlap texts are firstly separately tagged for the different representational meanings in the video. Based on the previous works~\cite{schroth2011exploiting,lei2020tvr}, this favors other tasks theoretically, such as video understanding, video retrieval, and video text translation.

(3) We first propose a new video text spotting framework with Transformer, termed \textbf{TransVTSpotter}, which solves the video multi-orient text spotting with a simple, but effective pipeline based on the tracking query-key mechanism and rotated boxes angle prediction.

(4) We evaluate and compare TransVTSpotter and other techniques for scene text detection, recognition, text tracking, and end-to-end video text spotting on BOVText and other existing datasets. Besides, a thorough analysis of performance on the proposed dataset is provided.

\section{Related Work}
\subsection{End-to-End Text Spotting}
For image-level text spotting, various methods~\cite{li2017towards, he2018end,lyu2018mask} based on deep learning have been proposed and have improved the performance considerably. Li \textit{et al.}~\cite{li2017towards} proposed the first end-to-end trainable scene text spotting method. The method successfully uses a RoI Pooling~\cite{ren2015faster} to joint detection and recognition features. Liao \textit{et al.}~\cite{lyu2018mask} propose a Mask TextSpotter which subtly refines Mask R-CNN and uses character-level supervision to detect and recognize characters simultaneously. 

However, these methods based on the static image can not obtain temporal information in the video, which is essential for some downstream tasks such as video understanding.
Compared to text spotting in a static image, video text spotting methods are rare. Yin \textit{et al.}~\cite{yin2016text} provides a detailed survey, summarizes text detection, tracking and recognition methods in video. 
Wang \textit{et al.}~\cite{wang2017end} introduced an end-to-end video text recognition method through associations of texts in the current frame and several previous frames to obtain final results. 
Cheng \textit{et al.}~\cite{cheng2019you} propose a video text spotting framework by only recognizing the localized text one-time. 
Nguyen \textit{et al.}~\cite{nguyen2014video} improves detection and recognition performance by temporal redundancy and linearly interpolate to recover missing detection results.
Rong \textit{et al.}~\cite{rong2014scene} tracked video text using tracking-by-detection. An MSER detector was used to locate scene text character, which was used as a constraint to optimize the trajectory search.
To promote video text spotting, we attempt to establish a standardized benchmark~(BOVText), covering various open scenarios and bilingual text annotation.

\subsection{Text Spotting Datasets for Images and Videos}
The various and practical benchmark datasets~\cite{karatzas2015icdar,veit2016coco,karatzas2013icdar,totaltext,karatzas2013icdar} contribute to the huge success of scene text detection and recognition at the image level. ICDAR2015~\cite{karatzas2015icdar} was provided from the ICDAR2015 Robust Reading Competition. Google glasses capture these images without taking care of position, so text in the scene can be in arbitrary orientations. ICDAR2017MLT~\cite{nayef2017icdar2017} is a large-scale bilingual text dataset, which is composed of complete scene images which come from 9 languages.
The COCO-Text dataset~\cite{veit2016coco} is currently the largest dataset for scene text detection and recognition. It contains 50,000+ images for training and testing.

The development of video text spotting is limited in recent years due to the lack of efficient data sets. ICDAR 2015 Video~\cite{karatzas2013icdar} consists of 28 videos lasting from 10 seconds to 1 minute in indoors or outdoors scenarios. Limited videos~(\ie{13 videos}) used for training and 15 for testing. Minetto Dataset~\cite{minetto2011snoopertrack} consists of 5 videos in outdoor scenes. The frame size is 640 x 480 and all videos are used for testing. YVT~\cite{nguyen2014video} contains 30 videos, 15 for training and 15 for testing. Different from the above two datasets, it contains web videos except for scene videos. USTB-VidTEXT~\cite{yang2017tracking} with only five videos mostly contain born-digital text sourced from Youtube. RoadText-1K~\cite{reddy2020roadtext} provides a driving videos dataset with 1000 videos. The 10-second long video clips in the dataset are sampled
from BDD100K~\cite{yu2018bdd100k}. As shown in Table.~\ref{table1}, the existing datasets contain a limited training set and single video scenarios. To promote the development of video text spotting, we create a large-scale, bilingual open-world benchmark dataset.

\subsection{Transformers in Vision}
Transformer is first proposed in~\cite{vaswani2017attention} as a new paradigm for machine translation. But recently, there is a popularity of using transformer architecture in vision tasks, such as detection~\cite{carion2020end,zhu2020deformable}, segmentation~\cite{zheng2021rethinking}, 3D data processing~\cite{zhao2020point}, video object tracking~\cite{wang2021transformer} and even backbone construction~\cite{dosovitskiy2020image}. Lately, some works~\cite{wang2021end,zeng2021motr,wang2021transformer} show using a transformer in processing sequential visual data also make remarkable shots. MOTR~\cite{zeng2021motr} introduce the concept of track query and the contiguous query passing mechanism for multiple-object Tracking. VisTR~\cite{wang2021end} solves instance segmentation by learning the pixel-level similarity and instance tracking is to learn the similarity between instances. 
But for video text spotting or multi-orient text tracking, to the best of our knowledge, there are still no transformer-based solutions while it is intuitive for its good capacity in temporal processing. 
Here, we propose the TransVTSpotter method and provide an affirmative answer to that, which shows convincingly high performance on the popular benchmark.

\section{BOVText Benchmark}

\subsection{Data Collection and Annotation}
\begin{figure}[t]
\begin{center}
\includegraphics[width=1\textwidth]{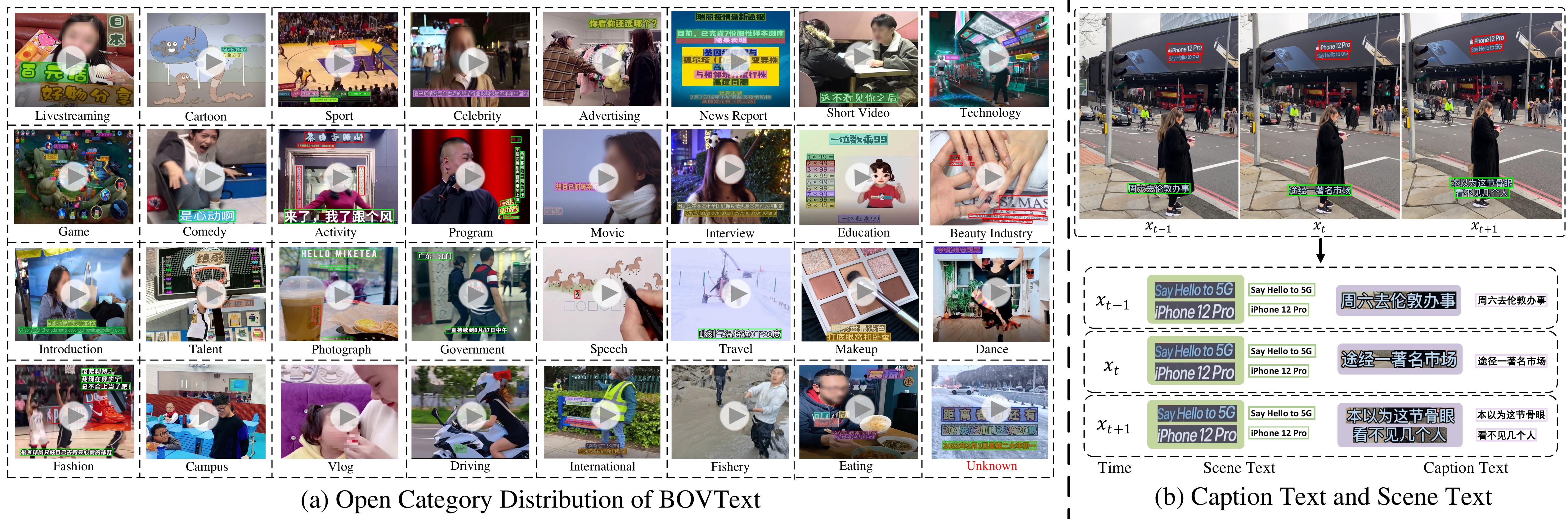}
\vspace{-5mm}
\caption{\textbf{Distributions of BOVText}. (a) The benchmark dataset covers a wide and open range of life scenes~(32 open-domain categories). (b) Caption text~(blue box) and scene text~(red box) are distinguished in BOVText, which is favorable for downstream tasks.}
\vspace{-4mm}
\label{fig2}
\end{center}
\end{figure}

\textbf{Data Collection.} To obtain abundant videos with various text types, we first start by acquiring a large list of different scenarios with text~(\eg{} \emph{game scenario}, \emph{travel scenario}) using  \textit{YouTube}\footnote{https://www.youtube.com/} and \textit{KuaiShou}\footnote{https://www.kuaishou.com/en} - an online resource that contains billions of videos with various scene text from cartoon movies to human relation.
Then, we choose 31 open-domain categories with 1 unknown category, \ie, \emph{Game}, \emph{Home}, \emph{Fashion}, and \emph{Technology}, as shown in Figure.~\ref{fig2}~(a). With each raw video category, we first choose the video clips with text, then make two rounds of manual screening to remove the ordinary videos without scene text and caption text. As a result, we obtain $2,021$ videos with $1,620,305$ video frames, as shown in Table~\ref{table1}.
Finally, to fair evaluation, we divide the dataset into two parts: the training set with $1,328,575$ frames from $1,541$ videos, and the testing set with $429,023$ frames from $480$ videos.
As shown in Figure~\ref{fig2}, different from the existing data sets, our dataset not only cares about scene text spotting in the real world, but also focuses on caption texts in the video. 
For the most part, caption text represents more global information than scene text, which is quite favorable for some downstream tasks, \eg{\textit{video understanding, video caption translation, etc.}}

\textbf{Data Annotation.}
We invite a professional annotation team to label each video text with four kinds of description information: the rotated bounding box describing the location information, judging the tracking identification(ID) of the same text, identifying the content of the text information, and distinguishing the category of the caption, title or scene text.
To save the annotation cost, we first sample the videos, annotate each sampled video frame, and then transform the annotation information from the sampled video frame to the unlabeled video frame by interpolation. Finally, we invite an audit team to carry out another round of annotation checks, and re-label part video frames with unqualified annotation.
\emph{For video sampling}, we use uniform sampling with a sampling frequency of $3$ to sample all the videos in the dataset, and obtain the sampled video frame set.
\emph{For sampling video frame annotation}, each text instance is labeled in the same quadrilateral way as in the ICDAR2015~\cite{zhou2015icdar}. In addition, the text instance also will be marked with two description information: the category of the caption, title or scene, the recognition content, and the tracking ID.
\emph{For interpolation on unlabeled video frames}, each text instance is marked with tracking ID and recognition content, so we can judge whether the texts in adjacent sampling frames are the same text with the same ID. For the same text instance, we first determine whether the text annotation of the sampled video frame is the starting and end frame of the text instance. If not, we look forward and backward for the starting and end position of the text instance and label it. Then we use the linear interpolation way to calculate the position of the text object in the middle of the unmarked video frame, and give tracking ID, recognition content, and category. 
\emph{For check and re-label bad cases,} the linear interpolation shows a dissatisfied performance in some cases, \emph{e.g., the new text appears on starting frame, text suddenly disappears on ending frame,} which are difficult to capture. Therefore, we invite an audit team to carry out another round of annotation checks. Around $150,000$ video frames with unqualified annotation from $1,670,305$ video frames are selected to refine, taking 20 men in three weeks.
As a labor-intensive job, the whole labeling process takes \textbf{30} men in three months, \ie{} \textbf{21,600} man-hours, to complete about \textbf{600,000} sampled video frame annotations.

\subsection{Dataset Analysis}
The statistic comparison between BOVText and other datasets are visualized in Figure~\ref{comparison} (a), \href{https://www.youtube.com/watch?v=HR5ZVZQWd3c}{\color{blue}{$\tt YouTube$}}, and summarized in Table~\ref{table1}. Besides, we provide more detailed information, such as `data distribution for 32 open scenarios', `text language and category distribution'.

To provide the community with unified text-level quantitative descriptions, we will compare our dataset with the previous datasets from four aspects, \ie text description, video scene, dataset size, and supported tasks.
\emph{For text description attribute} (\ie \emph{Category, MLingual}), our BOVText supports four types of text annotations(caption, title, scene, and other text) of video text with multi-language, which obviously has more extensive description ability than the existing dataset.

\begin{table}[t]
    \centering
	\caption{\textbf{Statistical Comparison.} `D',`R', `T', `S' and `BI' denotes the Detection, Recognition, Tracking, Spotting and bi-lingual text, respectively. `Incidental' denotes indoor and outdoor scenarios in daily life (\emph{e.g., walking outdoors, driving)}. `Open' refers to any scenarios, \emph{\eg{Game, Sport(NBA).}} \textcolor[RGB]{96,177,87}{In green} refers to these scenarios only supported by BOVText.}
	\label{table1}
	\vspace{-2mm} 
	\def\x{{$\footnotesize \times$}}
\scriptsize  
\setlength{\tabcolsep}{2.0pt}
\begin{tabular}{l|p{0.07\columnwidth}|c|c|c|c|c|p{0.4\columnwidth}}
    \whline
	Dataset &Category & BI & Task & Videos & Frames & Texts & Supported Scenario   \cr\shline \hline
	\hline
    AcTiV-D\cite{DBLP:conf/das/ZayeneSTHIA16}& Caption & - & D & 8 & 1,843 & 5,133 & News video\\
    UCAS-STLData\cite{DBLP:journals/mta/CaiWHML18} & Caption & - & D & 3 & 57,070 & 41,195 & Teleplay \\
    USTB-VidTEXT\cite{yang2017tracking}& Caption & - & D\&S & 5 & 27,670 & 41,932 &  Web video\\
    \hline
    YVT\cite{nguyen2014video} &Scene, Caption & - &  D\&R\&T\&S & 30 & 13,500 & 16,620 & Incidental:\tiny~Cartoon, Outdoor(supermarket, shopping street, driving...) 
    \cr\cline{8-8} 
    ICDAR2015 VT\cite{zhou2015icdar}&Scene & - &  D\&R\&T\&S & 51 & 27,824 & 143,588 & Incidental:\tiny~Outdoor(walking, driving, supermarket, shopping street...)
    \cr\cline{8-8} 
    LSVTD\cite{cheng2019you}&Scene & \checkmark &  D\&R\&T\&S & 100 & 66,700 & 569,300 & Incidental:\tiny~Indoor(shopping mall, supermarket, hotel...), Outdoor(driving...)
    \cr\cline{8-8} 
    RoadText-1K\cite{reddy2020roadtext}&Scene & - &  D\&R\&T\&S & 1,000 & 300,000 & 1,280,613 & Driving\\
    \hline
    BOVText(ours) & Scene, Caption, Title, Others & \checkmark & D\&R\&T\&S & \textbf{2,021} & \textbf{1,757,598} & \textbf{7,292,261} &   Open:\tiny~ Cartoon,~Vlog(supermarket, shopping street, driving), Travel (indoor and outdoor), \textcolor[RGB]{96,177,87}{Game(PUBG mobile...), Sport(NBA, world cup...), News , TV program, Education(campus, classroom, book...),Technology(introductory video, scientific propaganda...)...}\\
    
     \whline
\end{tabular}

	\vspace{-2mm} 
\end{table}

\emph{For video scene attribute} (\ie \emph{Scenario}), we present the 31 open scenarios and an "Unknown" scenarios distribution on BOVText in three levels
, \ie{} video, video frames, and text instances
, as shown in Table~\ref{supervision}.
%
%
Unlike the existing datasets, BOVText spans various video domains with these scenarios in the existing datasets~(\eg{Driving for RoadText-1k~\cite{reddy2020roadtext}, Vlog(supermarket, shopping street, indoor), Travel(hotel, railway station) for LSVTD~\cite{cheng2019you}}) and more open domains that are not yet supported~(\eg{Game(PUBG mobile, Honor of Kings...), Sport(NBA, world cup...), News}).
\emph{For the size of the dataset}(\ie \emph{Videos, Frames, Texts}), BOVText is \textbf{25} times larger than the existing largest dataset (\ie{} LSVTD~\cite{cheng2019you}) with various scenarios($1,757,598$ v.s $66,700$ video frames). RoadText1k~\cite{reddy2020roadtext} contains 300k videos frames, but the supported scenario is too single for only supports driving video scenarios.
\emph{For the supported tasks}, the proposed BOVText supports all video text tasks: detection, recognition, video text tracking, and end to end video text spotting.
For comprehensive research, we not only focus the scale, location, recognition content, and tracking ID, but also additionally collect and annotate the category of caption, title, scene or other texts for each text instance. 
As shown in Figure.~\ref{fig2}~(b), in a video, different types of text instances may exist simultaneously, and they are helpful to understand videos synergistically. 
Caption text can directly show the dialogue between people in video scenes and represent the time or topic of the video scenes, scene text can unambiguously define the object and can identify important localization and road paths in video scenes. 
Therefore, the text category annotation is favoring downstream tasks~(\eg{video text translation, video understanding, and video retrieval}), more details in the appendix.
In conclusion, the high efficiency of BOVText for evaluating advanced deep learning methods is very favorable for promoting various text spotting applications in real life.
\textbf{Statistic Comparison.} As shown in Table.~\ref{table1}, \emph{Category} denotes the category of the text type in the corresponding dataset. \emph{MLingual} denotes whether the dataset contains multiple language texts. \emph{Scenario} denotes the scene range of the video. \emph{Videos, Frames, Texts} represents the number of videos, video frames, video texts in the dataset, respectively. \emph{Task} denotes which tasks the dataset supports. 
%

\begin{table*}[t]
    \centering
    \vspace{-1mm}
    \caption{\textbf{The Data Distribution for 32 Open Scenarios}. \textcolor[RGB]{96,177,87}{In green} refers to these scenarios only supported by BOVText. "\%" denotes the percentage of each scenario data for whole data.}
    \def\x{{$\scriptsize \times$}}
\scriptsize
\setlength{\tabcolsep}{1.8mm}
\begin{tabular}{c|c|c|c|c|c|c|c}
    \whline
	Scenarios & Video & Video Frames & Text Instances & Scenarios & Video & Video Frames & Text Instances\cr\shline \hline
	\hline
	Cartoon & 67(3.2\%) & 64,359(3.6\%) & 123,191(2.1\%) & \textcolor[RGB]{96,177,87}{Sport} & 71(4.8\%) & 54,643(3.1\%) & 266,996(4.6\%)\\
	
	Vlog & 90(4.2\%) & 83,891(4.7\%) & 214,910(3.7\%) & \textcolor[RGB]{96,177,87}{News Report} & 100(4.1\%) & 66,207(3.7\%) & 178,000(3.1\%)\\
	
	Driving & 71(3.8\%) & 61,626(3.5\%) & 151,994(2.6\%) &
	\textcolor[RGB]{96,177,87}{Celebrity} & 50(2.4\%) & 39,958(2.3\%) & 121,235(2.1\%)\\
	
	Advertising & 32(1.7\%) & 28,645(1.6\%) & 91,090(1.0\%) & \textcolor[RGB]{96,177,87}{Technology} & 68(3.1\%) & 53,305(3.1\%) & 140,172(2.4\%)\\
	
	Activity & 35(1.6\%) &26,837(1.4\%) & 67,879(1.2\%) & \textcolor[RGB]{96,177,87}{Program} & 
	42(2.3\%)& 38,108(2.2\%)&214,561(3.7\%)
    \\
    
    Comedy & 88(4.5\%) & 79,206(4.5\%) & 317,865(5.5\%) & \textcolor[RGB]{96,177,87}{Game} & 21(1.0\%)&33,925(1.9\%)&84,106(1.5\%)\\
    
    Interview& 37(1.3\%) & 31,440(1.8\%) & 63,616(1.1\%) & \textcolor[RGB]{96,177,87}{Livestreaming}
    &64(3.1\%)&62,130(3.6\%)&211,569(3.6\%)
    \\
    
    Government & 66(2.5\%) & 45,457(2.6\%)& 93,874(1.6\%) & \textcolor[RGB]{96,177,87}{Speech} & 69(3.2\%) & 56,646(3.1\%) & 175,119(3.0\%)\\
    
    Travel &74(4.3\%)&71,291(4.1\%)&280,446(4.8\%) & \textcolor[RGB]{96,177,87}{Movie} &106(5.6\%)&108,110(6.3\%)&299,760(5.2\%)\\
    
    Campus & 52(2.3\%)&43,469(2.5\%)&139,760(2.4\%)&
    \textcolor[RGB]{96,177,87}{Photograph}&70(2.8\%)&64,025(3.6\%)&173,832(3.0\%)\\
    
    International&55(3.1\%)&52,774(3.6\%)&132,117(2.3\%) & \textcolor[RGB]{96,177,87}{Education} & 74(3.4\%)&59,824(3.6\%)&360,774(6.2\%)
    \\
    
    Short Video& 70(4.4\%)&59,756(3.4\%)&326,930(5.6\%)&
    \textcolor[RGB]{96,177,87}{Dance} & 43(1.9\%)&27,941(1.6\%)&71,740(1.2\%)\\
    
    \textcolor[RGB]{96,177,87}{Makeup}&63(3.1\%)&54,643(3.1\%)&111,814(1.9\%)&\textcolor[RGB]{96,177,87}{Fishery} & 81(4.5\%) &75,018(4.3\%)&230,085(4.0\%)\\
    
    \textcolor[RGB]{96,177,87}{Talent}&86(4.1\%)&71,024(4.1\%)&339,382(5.9\%)&\textcolor[RGB]{96,177,87}{Fashion}&63(3.0\%)&46,337(2.6\%)&98,942(1.7\%)
    \\
    
    \textcolor[RGB]{96,177,87}{Beauty Industry}&40(1.9\%)&41,350(2.4\%)&132,025(2.3\%)&
    \textcolor[RGB]{96,177,87}{Introduction}&64(3.8\%)&59,048(3.4\%)&236,721(4.1\%)\\
    
    \textcolor[RGB]{96,177,87}{Eating}&56(2.7\%)&62,893(3.6\%)&191,035(3.3\%)&
    \textcolor[RGB]{96,177,87}{Unknown}&53(2.5\%)&33,712(1.9\%)&150,721(2.6\%)\\
    
     \whline
\end{tabular}

	\vspace{-2mm} 
	\label{supervision}
\end{table*}

\subsection{Supported Tasks and Metrics} 
The proposed BOVText supports four tasks: (1)~Video Text Detection. (2)~Video Text Recognition. (3)~Video Text Tracking. (4)~End to End Text Spotting in Videos.

Following ICDAR2015~\cite{zhou2015icdar}~\footnote{https://rrc.cvc.uab.es/?ch=4\&com=tasks}, the evaluation protocols~\cite{wang2011end} are used for text detection and recognition task. 
For video text tracking and spotting task, the existing video text datasets such as ICDAR2015 (video)~\cite{karatzas2013icdar}~\footnote{https://rrc.cvc.uab.es/?ch=3\&com=tasks} and RoadText-1k~\cite{reddy2020roadtext} all adopted the MOT metrics~(\ie{}Multiple Object Tracking Accuracy ($MOTA$) and Multiple Object Tracking Precision ($MOTP$)). 
However, there are two sets of measures for Multiple Object Tracking: the MOT metrics~($MOTA$,$MOTP$)~\cite{bernardin2008evaluating} and ID metrics~($ID_{F1}$)~\cite{li2009learning,ristani2016performance}. 
The CVPR19 MOTChallenge evaluation framework~\cite{dendorfer2019cvpr19} presents that different measures serve different purposes. $Event\mbox{-}based$ measures like MOT help pinpoint the source of some errors and are thereby informative for the designer of certain system components.
$Identity\mbox{-}based$ measure($ID_{F1}$) is more favorable for evaluating how well computed identities conform to true identities. Except for using MOTA, MOTP, $Identity\mbox{-}based$ measures($ID_{F1}$), as a new metric is adopted firstly for video text spotting task. More detailed information for metric can be obtained in the appendix. 

\subsection{Our Method: TransVTSpotter}
Two ingredients are essential for direct text spotting for TransVTSpotter: (1) A set prediction loss that forces unique matching between predicted and ground truth multi-orient boxes. (2) An architecture that predicts a set of objects and associates the same objects during different frames.

\textbf{Multi-orient Box Matching.}
Compare to DETR~\cite{carion2020end}, 
the difference is that we propose an angle prediction and corresponding loss while only horizontal boxes prediction for DETR. Let us denote the ground truth set of objects by $y$, and $\hy = \{\hy_i\}_{i=1}^{N}$ the set of $N$ predictions. $y$ is as a set of size $N$ padded with $\noobject$ (no object).
To find a bipartite matching between these two sets we search for a permutation of $N$ elements $\sigma \in \Sigma_N$ with the lowest cost:
\vspace{-0.5mm}
\begin{equation}
\scriptsize
\label{eq:matching}
    \hat{\sigma} = \argmin_{\sigma\in\Sigma_N} \sum_{i}^{N} \lmatch{y_i, \hy_{\sigma(i)}},
\end{equation}
\vspace{-0.5mm}
where $\lmatch{y_i, \hy_{\sigma(i)}}$ is a pair-wise \emph{matching cost} between ground truth $y_i$ and a prediction with index $\sigma(i)$. 
The matching cost takes into account the class prediction, boxes prediction and the boxes rotated angle prediction.
Each element $i$ of the ground truth set can be seen as a $y_i = (c_i, b_i, a_i)$ where $c_i$ is the target class label, $b_i \in [0, 1]^4$ is a vector that defines ground truth box center coordinates and its height and width relative to the image size, and $a_i$ is rotation angle between the longest edge of ground truth multi-orient box and horizontal line~(x-axis). 
For the prediction with index $\sigma(i)$ we define probability of class $c_i$ as $\hp_{\sigma(i)}(c_i)$, the predicted box as $\hb_{\sigma(i)}$, and the predicted angle as $\ha_{\sigma(i)}$. Thus we define
$\lmatch{y_i, \hy_{\sigma(i)}}$ as $-\indic{c_i\neq\noobject}\hp_{\sigma(i)}(c_i) + \indic{c_i\neq\noobject} \bloss{b_{i}, \hb_{\sigma(i)}}+ \indic{c_i\neq\noobject} \aloss{a_{i}, \ha_{\sigma(i)}}$. Finally, we could compute the loss function with all pairs matched:
\vspace{-0.5mm}
\begin{equation}
\scriptsize
\hloss{y, \hy} = \sum_{i=1}^N \left[-\log  \hp_{\hat{\sigma}(i)}(c_{i}) + \indic{c_i\neq\noobject} \bloss{b_{i}, \hb_{\hat{\sigma}}(i)+\indic{c_i\neq\noobject} \aloss{a_{i}, \ha_{\sigma(i)}}}\right]\,,
\end{equation}
\vspace{-0.5mm}
where $\bloss{\cdot}$ a linear combination of the $\ell_1$ loss and the generalized IoU loss~\cite{rezatofighi2019generalized,carion2020end}. And $\aloss{\cdot}$ refers to a cosine embedding loss  with $1-cos(\ha_{\sigma(i)}-a_{i})$. 


\begin{figure}[t]
\begin{center}
\includegraphics[width=1\textwidth]{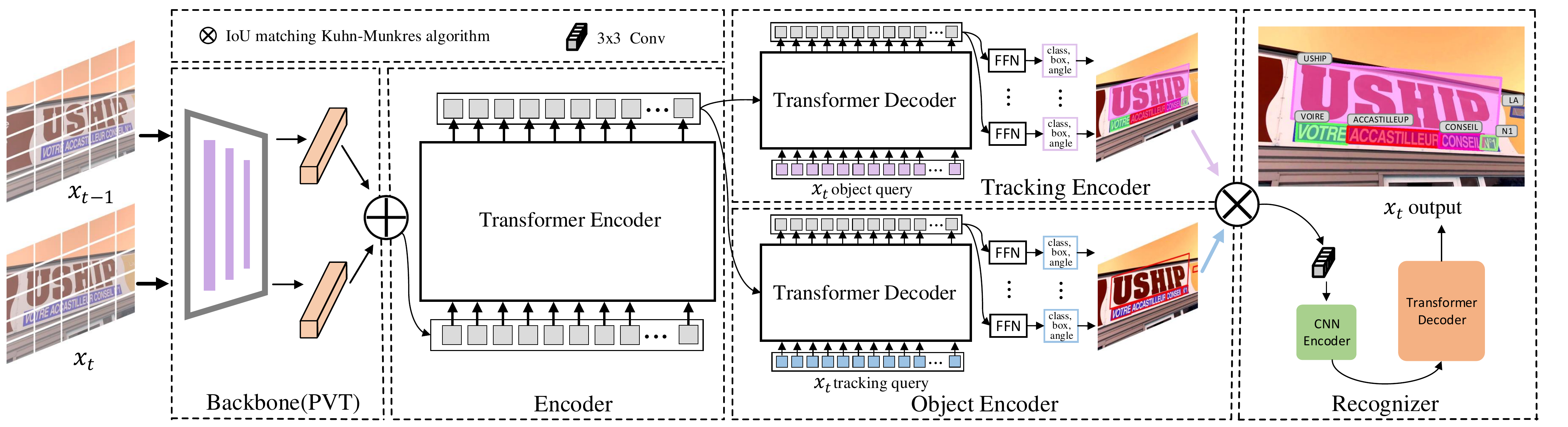}
\vspace{-5mm}
\caption{\textbf{ Pipeline of TransVTSpotter}. It contains four main components: 1) A transformer backbone(PVT~\cite{wang2021pyramid}) extracts feature representation of multiple images; 2) A transformer encoder models the relations of pixel-level features; 3) Two transformer decoders~(shared weight) decode the instance-level features; 4) Attention-based recognizer~\cite{lu2021master} recognizes each text instance.}
\label{pipeline}
\end{center}
\vspace{-4mm}
\end{figure}
\textbf{TransVTSpotter architecture.}
The overall TransVTSpotter architecture is surprisingly simple and depicted in Figure.~\ref{pipeline}. A transformer-based backbone~\cite{wang2021pyramid} is used to extract feature representation, transformer-based encoder-decoder framework learned current object query and previous frame tracking query as input and predicts \textit{detection boxes} and \textit{tracking boxes}~\cite{sun2020transtrack}. With the detection boxes and tracking boxes, box IoU matching is used to obtain the final tracking result. Finally, attention-based recognizer~\cite{lu2021master} is utilized to obtain the final recognition results. \textbf{Text Tracking with Transformer.} 
Text Tracking with Transformer includes three components: backbone, transformer encoder and transformer decoder, as shown in Figure.~\ref{pipeline}. 
\textit{Backbone}. Starting from the two consecutive frames $x_{\rm t} \in\Re^{3\times H_0\times W_0}$ and $x_{\rm t\mbox{-}1} \in\Re^{3\times H_0\times W_0}$, a transformer
backbone~\cite{wang2021pyramid} generates a lower-resolution activation map for the two frames($f_{\rm t} \in\Re^{C\times H\times W}$ and $f_{\rm t\mbox{-}1} \in\Re^{C\times H\times W}$), then a new feature sequence$f_{\rm t}^*$ can be obtained by simple concatenating $f_{\rm t}$ and $f_{\rm t\mbox{-}1}$. The extracted features $f_{\rm t}$ of the current frame are temporarily saved and then re-used for the next frame. 
\textit{Transformer Encoder.} We adopted deformable transformer encoder~\cite{zhu2020deformable} to model the similarities among all the pixel level features for the extracted features $f_{\rm t}$.
\textit{Transformer Decoder.} Two parallel decoders~\cite{sun2020transtrack} are employed. 
The object decoder takes learned object query~\cite{carion2020end} as input and predicts detection rotated boxes. The tracking decoder takes the object feature from previous frames as input and predicts the corresponding tracking rotated boxes. Finally, with detection rotated boxes and tracking rotated boxes, TransTT obtains the final tracking result by box IoU matching and the Kuhn-Munkres(KM) algorithm~\cite{kuhn1955hungarian}. \textbf{Recognizer.} Following MASTER~\cite{lu2021master} is utilized to predict output sequence with 2D-attention.

\section{Experimental}
In this section, we mainly conduct experiments on our BOVText. More experiments, such as the performance of TransVTSpotter in other datasets, would be provided in the appendix.

\subsection{Implementation Details}
Except for the TransVTSpotter, we also adopt CRNN~\cite{shi2016end}, RARE~\cite{shi2016robust} as the recognition baseline and PSENet~\cite{psenet}, EAST~\cite{zhou2017east}, DB~\cite{liao2020real} as the detection baseline to evaluate our BOVText.
\textit{Detection}: we train detectors with pre-trained model on COCOText~\cite{veit2016coco}. \textit{Recognition}: the network is pre-trained on the \textit{chinese ocr}\footnote{https://github.com/YCG09/chinese\_ocr} and MJSynth~\cite{jaderberg2014synthetic}, then fine-tuned on our BOVText.
All of our experiments are conducted on 8 V100 GPUs.
%
%
AdamW~\cite{loshchilov2017decoupled} as the optimizer and the batch size is set to be 16. The initial learning rate is 2e-4 for the transformer and 2e-5 for the backbone. The weight decay is 1e-4 All transformer weights are initialized with Xavier-init~\cite{glorot2010understanding}. The data augmentation includes random horizontal, scale augmentation, resizing the input images whose shorter side is by 480-800 pixels while the longer side is by at most 1333 pixels. The model is first pre-trained on COCOText~\cite{veit2016coco} and then fine-tuned on other video text training sets. For each iteration, two adjacent frames are randomly selected from 
one video from training set to train our model.

\subsection{Attribute Experiments Analysis for BOVText}

\textbf{New Scenarios, New Challenge for Video Text Tasks.}
Figure.~\ref{fig5} and Table.~\ref{table7} gives the tracking performance $ID_{F1}$ of TransVTSpotter in different scenarios of BOVText. Two new insights can be present from the figure and table: 1) The existing benchmark datasets cannot effectively test the effectiveness of advancing algorithm on some novel scenarios~(\eg{} NewsReport, Cartoon) for first proposed in BOVText.
LSVTD~\cite{cheng2019you} and RoadText-1k~\cite{reddy2020roadtext}, as the two largest data sets on the existing video text datasets are used to compare with our BOVText. TransVTSpotter achieves a tracking performance ${\rm ID_{F1}}$ of $70.4\%$ on $NewsReport$ scenario with BOVText training set, around $20$ percent point improvement than training with LSVTD~\cite{cheng2019you} and RoadText-1k~\cite{reddy2020roadtext}. We argue that the main cause is that there existing a mass of caption texts in $NewsReport$ scenario, but LSVTD and RoadText almost no such dense text scenario, which is a new challenge for algorithms.
%
%
Besides, training with only RoadText-1k obtains a low performance no matter which scenarios. There are two main causes for this. Firstly, the location annotation of RoadText-1k is an upright bounding box(two points), but the counterpart of BOVText is multi-orient boxes(four points). Secondly, the data domain is entirely different for the two datasets. Compare with various scenarios~(\eg{} Game, Sports) and text types~(\eg{} long caption text, big text), the scenario of RoadText-1k only contains small and low-resolution road signs, plate number on driving scenarios. 
2) Huge performance gap existing during different scenarios. 
As shown in Figure.~\ref{fig5}, the model achieves the best performance with a $ID_{F1}$ of $88.4\%$ in $Fishery$ videos, since the conspicuous text instances, simple foreground~(\ie{} caption texts) and background are designed in cartoon videos. By comparison, several scene categories obtain extremely dissatisfied performance due to complex background, various text appearance, and unsteady camera movements, such as $Sports$ of $46.7\%$ and $Travel$ of $55.8\%$.

\begin{figure}[t]
\begin{center}
\includegraphics[width=1\textwidth]{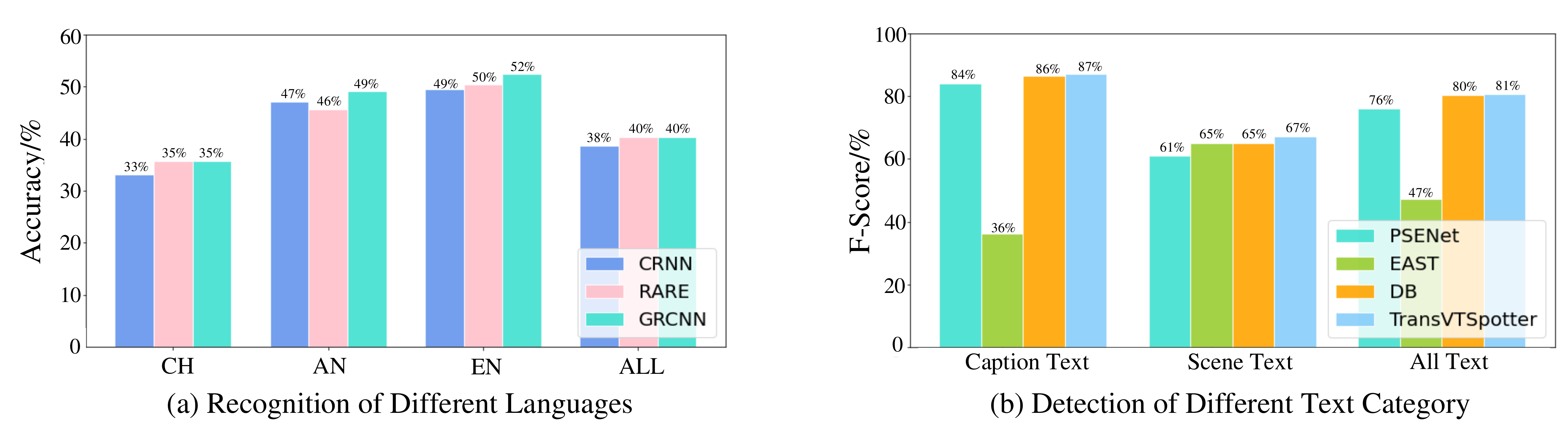}
\vspace{-5mm} 
\caption{\textbf{Attribute Experiments of BOVText}. (a) Recognition accuracy of different models in different languages. (b) Detection of different model in caption or scene text.  `CH', `AN', `EN' and `ALL' refer to `Chinese', `Alphanumeric', `English' and `All Characters', respectively.}
\label{fig3}
\end{center}
\vspace{-3mm} 
\end{figure}


\begin{figure}[t]
\begin{center}
\includegraphics[width=1\textwidth]{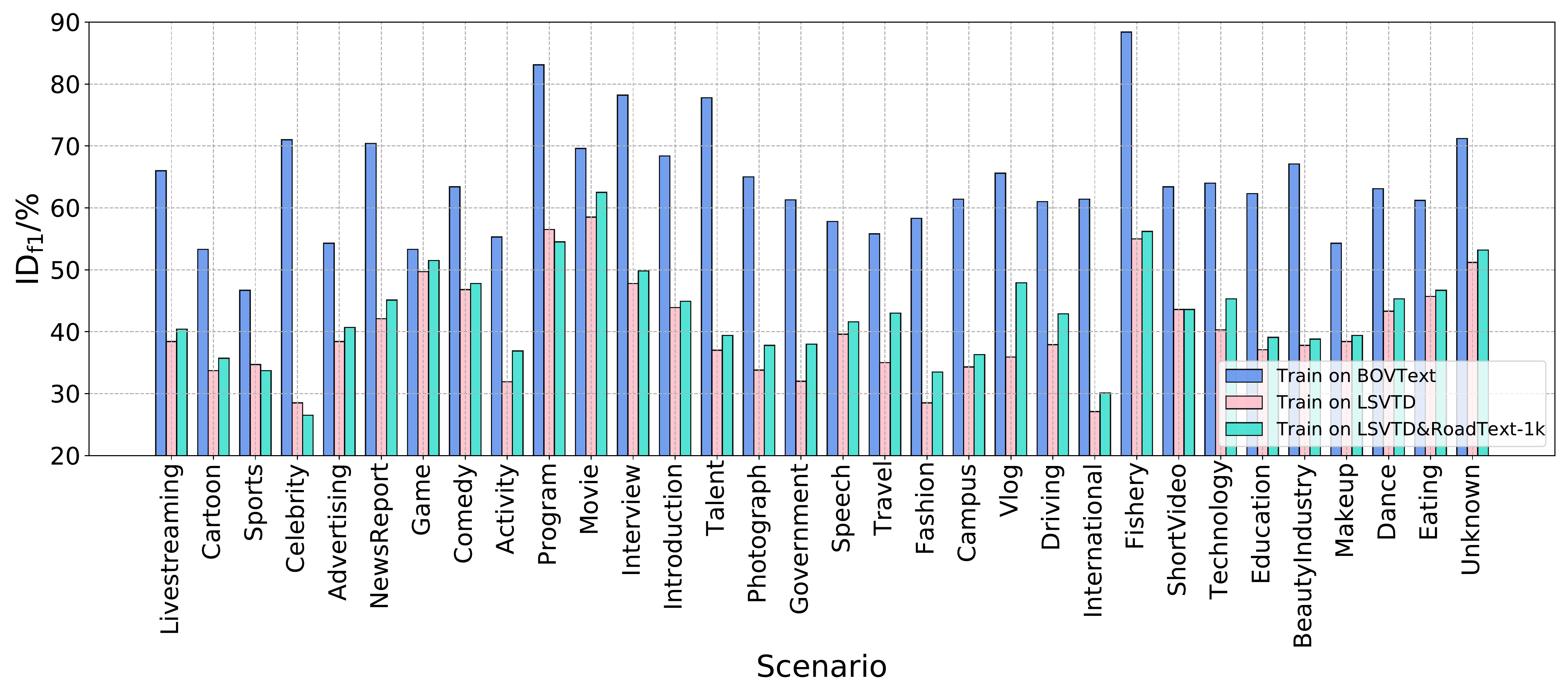}
\vspace{-7mm} 
\caption{\textbf{Tracking performance~(\ie{$ID_{F1}$}) with TransVTSpotter in different scenarios}. `Training on LSVTD' and `Training on BOVText' denotes training on LSVTD train set and BOVText train set, respectively. Huge performance gap existing during different scenarios.}
\label{fig5}
\end{center}
\vspace{-5mm} 
\end{figure}

\begin{table}[t]
    \centering
	\caption{\textbf{Attribute Experiments for Scenarios.} Random scenarios are selected to present here.}
	\label{table7}
	\vspace{-1mm} 
	\def\x{{$\footnotesize \times$}}
\scriptsize
\setlength{\tabcolsep}{2pt}
\begin{tabular}{c|c|ccccc|ccccc}
    \whline
	\multirow{2}*{$\rm Method$} &
	\multirow{2}*{\makecell[c]{Training Set}} &
	\multicolumn{5}{c|}{Detection (F-score/\%) on BOVText}&
	
	\multicolumn{5}{c}{Tracking (${\rm ID_{F1}}$/\%) on BOVText}   \cr\cline{3-12}  
	
	
	~ & ~ & Cartoon  & NewsReport & Driving & Program & Avg. & Cartoon  & NewsReport & Driving & Program & Avg. \cr\shline \hline
	\hline

	\multirow{4}*{TransVTSpotter} & LSVTD & 48.2  & 68.2 & 60.3 & 74.2& 63.8 & 33.7 & 42.1 & 37.9 & 56.5 & 38.1\\
	~ & RoadText & 5.6 & 4.3 & 3.0 & 9.6 & 6.7 & 0.7 & 3.2 & 2.5 & 4.2 & 3.5  \\
	~ & LSVTD\&RoadText & 45.9 & 70.4  & 62.3 & 75.3 & 67.6 & 35.7 & 50.5  & 40.9 & 54.5 & 41.3 \\
	~ & BOVText & \textbf{80.7 }& \textbf{82.3} & \textbf{78.1} & \textbf{91.3} & \textbf{81.7} & \textbf{53.3} & \textbf{70.4} & \textbf{61.0} & \textbf{83.1} & \textbf{64.7}\cr\hline 
     \whline 
\end{tabular} 

	\vspace{-3mm} 
\end{table}

\textbf{Bilingual Recognition, New Challenge.} 
As shown in Figure.~\ref{fig3}~(a), the text recognition results for different languages are provided. 
In summary, the alphanumeric recognition result~(about $47\%$) is better than the Chinese recognition result~(about $35\%$), regardless of the models. The final results~(about $40\%$) for all characters are satisfactory, can not meet the requirement of the application. Unlike English, Chinese contains thousands of characters($3,856$ Chinese characters v.s. $26$ English characters on BOVText), which are difficult to recognize.

\textbf{Long Caption Text, New Challenge.} As shown in Figure.~\ref{fig3}~(b), for DB~\cite{liao2020real}, PSENet~\cite{psenet} and our TransVTSpotter, the performance of caption text is better than the counterpart of scene text~(around $80\%$ vs. $60\%$) due to the more clear and bigger caption text. 
But for EAST~\cite{zhou2017east}, long caption text show a low performance with $36\%$ F-score. 
The prime reason is that caption texts are all long text ( average width-height ratio: $6.8$ for caption $v.s.$ $2.3$ for scene text on BOVText), a different case for EAST~\cite{zhou2017east}, as shown in \href{https://www.youtube.com/watch?v=CuLk0wYvqmU}{\color{blue}{$\tt YouTube Demo$}}.
However, the existing video text datasets hardly contain long caption texts, our BOVText can fill out the gap for a more comprehensive evaluation of text types.

\begin{table}[t]
    \centering
	\caption{\textbf{Detection and Recognition Performance on BOVText}.}
	\label{table4}
	\vspace{-2mm} 
    \def\x{{$\footnotesize \times$}}
\scriptsize
\setlength{\tabcolsep}{2.0pt}
\begin{tabular}{c|ccc|c|cccc|cccc}
    \whline
	\multicolumn{4}{c|}{Detection Performance/\%} & \multicolumn{9}{c}{ Recognition Performance/\%}  
	\cr\cline{0-12}  
	\multirow{2}*{Method} & \multirow{2}*{Precision} & \multirow{2}*{Recall} & \multirow{2}*{F-score} & \multirow{2}*{Method} & \multicolumn{4}{c|}{Pretrained} & \multicolumn{4}{c}{Fine tuned}
	\cr\cline{6-13}
	~ & ~ & ~ & ~ & ~ & Chinese & Alphanumeric & English & All & Chinese & Alphanumeric & English & All 
	\cr\shline \hline 
	\hline
	EAST~\cite{zhou2017east} & 55.4 & 40.8 & 47.0 & CRNN~\cite{shi2016end} & 26.0 & 32.1 & 36.1 & 23.2 & 33.2 & 47.1 & 49.5  & 38.6 \cr
	
	PSENet~\cite{psenet} & 78.3 & 75.7 & 77.0 & RARE~\cite{shi2016robust} & 25.2 & 34.2 & 37.4 & 23.5 & 35.6  & 45.7 & 50.4 & 40.2 
	\cr
	DB~\cite{liao2020real} & 84.3 & 77.6 & 80.8 & GRCNN~\cite{wang2017gated} & 23.1 & 39.8& 40.4 & 26.7 & 35.6 & 49.2 & 52.4  & 40.3 
	\cr
	TransVTSpotter & \textbf{86.2} & \textbf{77.4} & \textbf{81.7} & - & 26.2 & 40.3 & 42.1 & 29.1 & 36.2 & 48.9 & 52.1 & 40.4  
	
    \cr\hline

     \whline
\end{tabular}

	\vspace{-2mm} 
\end{table}

\begin{table}[t]
    \centering
	\caption{\textbf{Text Tracking and End to End Video Text Spotting Performance on BOVText.} Text tracking trajectory id generation use a method proposed in~\cite{wang2017end}.} 
	\label{table3}
	\vspace{-2mm} 
	\def\x{{$\footnotesize \times$}}
\scriptsize
\setlength{\tabcolsep}{2pt}
\begin{tabular}{c|c|ccc|cc|ccc|cc}
    \whline
	\multicolumn{2}{c|}{$\rm Method$} &
	\multicolumn{5}{c|}{Text Tracking on BOVText} & \multicolumn{5}{c}{End to End Text Spotting on BOVText} \cr\hline
	
	
	Detection & Recognition & $\rm ID_P$/\%& $\rm ID_R$/\% & ${\rm ID_{F1}}$/\% & $\rm MOTA$& $\rm MOTP$ & $\rm ID_P$/\% & $\rm ID_R$/\% & ${\rm ID_{F1}}$/\% & $\rm MOTA$& $\rm MOTP$ \cr\shline \hline
	\hline
	\multirow{2}*{EAST~\cite{zhou2017east}}
     & CRNN~\cite{shi2016end} & \multirow{2}*{29.9} & \multirow{2}*{26.5} & \multirow{2}*{28.1}  & \multirow{2}*{-0.216} & \multirow{2}*{0.758} & 6.8 & 6.9 & 6.8 & -0.793 & 0.763 \cr
     ~ & RARE~\cite{shi2016robust} & ~ & ~ & ~ & ~ & ~ & 4.2 & 5.3 & 4.7 & -1.05  & 0.772\cr\cline{1-12}

     \multirow{2}*{PSENet~\cite{psenet}}
     & CRNN~\cite{shi2016end}  & \multirow{2}*{52.4} & \multirow{2}*{40.9} & \multirow{2}*{45.9}  & \multirow{2}*{0.521} & \multirow{2}*{0.775} & 31.3 & 26.7 & 28.8 & -0.170 & 0.792 \cr
     ~ & RARE~\cite{shi2016robust}& ~ & ~ & ~ & ~ & ~ & 35.6 & 28.8 & 31.7 & -0.162 & 0.803 \cr\cline{1-12}

     \multirow{2}*{DB~\cite{liao2020real}}
     & CRNN~\cite{shi2016end} & \multirow{2}*{55.2} & \multirow{2}*{42.9} & \multirow{2}*{48.3} & \multirow{2}*{0.532} & \multirow{2}*{0.783} & 38.8 & 30.1 & 33.7 & -0.132 & 0.813 \cr  
     ~ & RARE~\cite{shi2016robust}& ~ & ~ & ~ & ~ & ~ & 41.1 & 29.3 & 34.2 & -0.126 & 0.811 \cr\cline{1-12}
     
     \multicolumn{2}{c|}{TransVTSpotter(ours)} & \textbf{71.0} & \textbf{59.7} & \textbf{64.7} & \textbf{0.682} & \textbf{0.821}  & \textbf{43.6}& \textbf{38.4} & \textbf{40.8} & \textbf{-0.014} & \textbf{0.820} \cr\cline{1-12}
     \whline
\end{tabular}

	\vspace{-3mm}  
\end{table}

\subsection{Text Detection, Recognition, Tracking and Spotting on BOVText}
\textbf{Video Text Detection and Recognition.}
As shown in Table.~\ref{table4}, image-based text detection on BOVText is not unsatisfactory,  with lower results than these methods report on existing text datasets. 
For example, EAST obtains an f-score of $47.0\%$ compared to the F-score of $80.7\%$ on icdar2015~\cite{zhou2015icdar}, but our TransVTSpotter obtain an f-score of $81.7\%$ on BOVText, at least $1\%$ improvement compare to the image-based detectors~(\ie{} DB, PSENet and EAST).
For text recognition, CRNN~\cite{shi2016end} based on CTC loss, RARE~\cite{shi2016robust} with attention mechanism and GRCNN~\cite{wang2017gated} as the base text recognizers to test our BOVText. 
The text annotation in our BOVText covers two languages~(\ie{English and Chinese}), thus we conduct several experiments for each language. 
The recognition model only yields about $40\%$ accuracy on our dataset, but the same model reports at least $90\%$ on most benchmark datasets~\cite{karatzas2013icdar} for text recognition. 
The main reasons have two points: (1) The proposed BOVText is bilingual, and the category number of Chinese characters in real-world is much larger than those of Latin languages. 
(2) The video texts are quite blurred, out-of-focus, and the distribution of characters is relatively smaller than the static image counterparts, which presents more challenges.  

\textbf{Video Text Tracking.} As shown in Table.~\ref{table3},
$\rm ID_{F1}$~($64.7\%$) of our TransVTSpotter achieves the best performance, at least $10\%+$ improvement than other methods. 
Besides, without NMS and other post-processing, TransVTSpotter presents 9 fps no matter which dataset.
More details and analysis concerning TransVTSpotter can be obtained in the appendix.
And EAST shows the worst performance with a $\rm ID_{F1}$ of $28.1\%$.
The $\rm IDF_1$ of EAST~\cite{zhou2017east} is lower with $17.8\%$ gap than that of PSENet~\cite{psenet}. The main reason is that BOVText contains a mass of long text instances, but regression-based EAST can not deal with the long text cases well. 
The performance of DB is similar to that of PSENet for both all are the segmentation-based methods.

\textbf{End to End Text Spotting in Video}. Detection and text tracking tasks are paving the way for the recognition task. Table.~\ref{table3} shows the performance of text spotting on BOVText. Similar to the text tracking performance, our TransVTSpotter achieves the state-of-the-art performance with at least $6\%$ $\rm ID_{F1}$ improvement compared to the other methods.
Besides, the MOTP of TransVTSpotter achieves $82\%$, $1\%$ percent points improvement than the counterpart of using DB and RARE. 
The great performance for $40.8\%$ $\rm ID_{F1}$ and $82\%$ MOTP present satisfactory tracking and recognition trajectory and detection results, respectively.
The corresponding performance using EAST~\cite{zhou2017east} as the detector in video text spotting is still not satisfied with around $5\%$ $\rm ID_{F1}$ and $-0.8$ MOTA. 
Without TransVTSpotter, the combination of DB~\cite{liao2020real} and RARE~\cite{shi2016robust} achieves the best performance with a $34.2\%$ $\rm ID_{F1}$, but there is at least $6\%$ gap compare to our method.

\subsection{Potential Negative Societal Impacts and Solution}
We argue that there mainly exits slight potential negative societal impacts for personal privacy. Although much personal information, \eg{\emph{names, identifying information, human faces}}, have been blurred to protect privacy, there still might exist a little risk. 

\textbf{How to blur human faces.} We blur the human faces in BOVText with four steps. Firstly, human faces in each frame would be detected by \textit{face recognition}\footnote{https://github.com/ageitgey/face\_recognition} - a powerful, simple, and easy-to-use face recognition open source project with complete development documents and application cases. Secondly, with the location box from the previous step, we extract face ROI from the original image. Thirdly, we blur the face ROI with Gaussian Blur operation in OpenCV\footnote{https://www.tutorialspoint.com/opencv/opencv\_gaussian\_blur.htm}. Finally, we store the blurred face in the original image and recover to video.

\subsection{Link to Other Video-and-Language Applications}

In this section, we show that the practicability of the proposed BOVText, not a toy benchmark, which can promote other video-and-text application research. Text spotting in static images has numerous application scenarios: (1) Automatic data entry. SF-Express~\footnote{https://www.sf-express.com/cn/sc/} utilizes OCR techniques to accelerate the data entry process. NEBO~\footnote{https://www.myscript.com/nebo/} performs instant transcription as the user writes down notes. (2) Autonomous vehicle~\cite{mammeri2014road,mammeri2016mser}. Text-embedded panels carry important information, \eg{geo-location, current traffic condition, navigation}, and etc.
(3) Text-based reading comprehension. TextCaps~\cite{sidorov2020textcaps} and text-based VQA~\cite{singh2019towards,biten2019scene} show the new vision-and-language tasks, which need to recognize text, relate it to its visual context, semantic, and visual reasoning between multiple text tokens and visual entities, such as objects. 
Similarly, there are many application demands for video text understanding across various industries and in our daily lives. We list the most outstanding ones that significantly impact, improving our productivity and life quality.
\textbf{Firstly}, automatically describing video with natural language~\cite{xu2016msr,wang2019vatex} can bridge video and language. 
\textbf{Secondly}, video text automatic translation~\footnote{https://translate.google.com/intl/en/about/} can be extremely helpful as people travel, and help video-sharing websites~\footnote{https://www.youtube.com/} to cut down language barriers. 
\textbf{Finally,} text-based video retrieval~\cite{lei2020tvr,liu2020violin} is a irreplaceable business for many companies, such as Google and YouTube.
More details and analyses for application scenarios concerning BOVText in the appendix.

\section{Conclusion and Future Work}
In this paper, we establish a large-scale, bilingual open-world benchmark dataset for video text tracking and spotting, termed BOVText, with four description information, \ie, bounding box, tracking ID, recognition content, and text category label. 
Compare with the existing benchmarks, the proposed BOVText mainly contains four advantages: large-scale training set~(\ie{2,021+video}), $32$ open real scenarios (\textit{Sportscast, Life Vlog, Game}), bilingual annotation, and abundant text types annotation(Caption, title, and scene text). 
Besides, we first propose an end-to-end video text spotting framework with Transformer, termed TransVTSpotter, which presents a simple, but efficient attention-based query-key pipeline.  
On ICDAR2015(video), TransVTSpotter achieves the state-of-the-art performance.
In general, we hope the proposed BOVText and TransVTSpotter would provide a standard benchmark to facilitate the advance of video-and-text research. 

\section{Appendix}

\subsection{BOVText Metrics}

The proposed BOVText includes four tasks: (1)~Video Text Detection; (2) Video Text Recognition; (3) Video Text Tracking. (4)~End to End Text Spotting in Videos. 
$MOTP$ (Multiple Object Tracking Precision)~\cite{bernardin2008evaluating}, $MOTA$ (Multiple Object Tracking Accuracy) and $IDF_1$~\cite{dendorfer2019cvpr19,ristani2016performance} as the three important metrics are used to evaluate task3~(text tracking) and task4~(text spotting) for BOVText. Following the previous works~\cite{karatzas2013icdar,reddy2020roadtext}, BOVText evaluates text tracking methods in video and compare their performance with the MOTA and MOTP, which are given by:

\begin{equation}
\scriptsize
MOTP = \frac{{\textstyle \sum_{i,t}^{}}(1-d^i_t)}{\textstyle \sum_{t}^{}c_t}\,,
\label{equation1}
\end{equation}

where $c_t$ denotes the number of matches found for time $t$. For each of these matches, calculate the iou $d^i_t$ between the object $o^i$ and its corresponding hypothesis. It shows the ability of the tracker to estimate precise object positions. MOTA is calculated as follows:

\begin{equation}
\scriptsize
MOTA = 1-\frac{{\textstyle \sum_{t}^{}}(m_t+fp_t+mme_t)}{\textstyle \sum_{t}^{}g_t}\,,
\label{equation2}
\end{equation}
where $m_t$, $fp_t$ and $mme_t$ are the number of misses, false positives, and mismatches, respectively. $g_t$ is the number of objects present at time $t$. It shows the tracker’s performance at detecting objects and keeping their trajectories, independent of the precision of the location. $ID_{F1}$ is the ratio of correctly identified detections over the average number of ground-truth and computed detections. And the metric is more reasonable to evaluate ID switches in some cases.
We evaluate the metrics in BOVText by:
\begin{equation}
\scriptsize
ID_{tp} =  \sum_{h}^{} \sum_{t}^{} m(h,o,\bigtriangleup_t,\bigtriangleup_s)\,,
\end{equation}

\begin{equation}
\scriptsize
ID_{F1} = \frac{2ID_{tp}}{2ID_{tp}+ID_{fp}+ID_{fn}}\,,
\end{equation}

where $ID_{tp}$, $ID_{fp}$ and $ID_{fn}$ refer to true positive, false positive and false negative of matching ID. Besides, the ID metric~\cite{dendorfer2019cvpr19} also includes $MT$~(Mostly Tracked) Number of objects tracked for at least 80 percent of lifespan, $ML$~(Mostly Lost) Number of objects tracked less than 20 percent of lifespan. $\bigtriangleup_t$ and $\bigtriangleup_s$ refer to time matching and space location matching, respectively.

For Task4~(End to End Text Spotting in Videos), the objective of this task is to recognize words in the video as well as localize them in terms of time and space.  We use $ID_{F1}$ to evaluate our BOVText, which focuses on text instance ID tracking and recognition results that be required by many downstream tasks. More specifically, 

\begin{equation}
\scriptsize
ID_{tp} =  \sum_{h}^{} \sum_{t}^{} m(h,o,\bigtriangleup_t,\bigtriangleup_s,\bigtriangleup_r )\,,
\end{equation}

\begin{equation}
\scriptsize
ID_{F1} = \frac{2TID_{tp}}{2TID_{tp}+TID_{fp}+TID_{fn}}\,, 
\label{equation6}
\end{equation}

where $\bigtriangleup_t$, $\bigtriangleup_s$ and $\bigtriangleup_r$ refer to ID matching, space location matching and recognition result matching. $h$ and $o$ denote hypothesis set~(\eg{predicted ID $I_p$ , box locations $L_p$, recognition results $R_p$}) and ground truth set with~(ID $I_g$, box locations~$L_g$, recognition ground true~$R_g$). And the three matching can be obatined by:

\vspace{-2mm}
\begin{eqnarray}
\scriptsize
\bigtriangleup_t: I_p=I_g ,              &          \bigtriangleup_s: IoU(L_p,L_g)>0 ,      &    \bigtriangleup_r:  R_p= R_g.
\label{equation7}
\end{eqnarray}

The match of $h$ and $o$ is a true positives of text ID~(\ie{$ID_{tp}$}) when these conditions~(\ie{$\bigtriangleup_t$, $\bigtriangleup_s$ and $\bigtriangleup_r$} are met. Similarly, false positive~(\ie{$ID_{fp}$}) and false negative~(\ie{$ID_{fn}$}) of text ID can be obtained for $ID_{F1}$ calculation.

\subsection{More details for experiments}
\textbf{TransVTSpotter.} All experiments are conducted on Tesla V100 GPU. We train the model for 150 epochs and the learning rate drops by a factor of 10 at the 100th epoch.

\textbf{Baselines concerning the existing methods.}
Video-based text spotting methods are rare and lack open-source algorithms. Therefore, we adopt various mature image-based techniques to compare and evaluate the efficiency of BOVText. \textbf{Detection}.
EAST~\cite{zhou2017east} as one of the popular regression-based methods is used to test our BOVText. The method adopts FCNs to predict shrinkable text score maps, rotation angles. 
For segmentation based methods, we adopt PSENet~\cite{psenet} and DB~\cite{liao2020real} to evaluate our BOVText. PSENet~\cite{psenet} generates various scales of shrinked text segmentation maps, then gradually expands kernels to generate the final text instance. \textbf{Recognition}.
Recent methods mainly include two techniques, Connectionist Temporal Classification~(CTC) and attention mechanism. 
In CTC-based methods, CRNN~\cite{shi2016end} as the representation, which introduced CTC decoder into scene text recognition with BiLSTM to model the feature sequence. 
In Attention-based methods, RARE~\cite{shi2016robust} normalizes the input text image using the Spatial Transformer Network~(STN~\cite{jaderberg2015spatial}). Then, it estimates the output character sequence from the identified features with the attention module. \textbf{Text Tracking Trajectory Generation}.
With text detection and recognition in a static image, we only obtain text localization and recognition information without temporal information, which are insufficient for video spotting evaluation~(\eg{$ID_{F1}$,$MOTA$ and $MOTP$}).
Following the work~\cite{wang2017end}, we link and match text objects in the current frame and several frames by IOU and edit the distance of text. 
All of the experiments use the same strategy: (1) Training detector and recognizer with BOVText. (2) Matching text objects with corresponding text tracking trajectory id.

\subsection{Ablation study for TransVTSpotter}

\begin{table}[t]
    \centering
	\caption{\textbf{Ablation study for angle prediction.} The gaps of at least ~(\green{+7.1\%}) improvement after using angle prediction are shown in green.}
	\label{table9}
	\vspace{-2mm} 
	\def\x{{$\footnotesize \times$}}
\scriptsize
\setlength{\tabcolsep}{2pt}
\begin{tabular}{c|ccc|ccc}
    \whline
	\multirow{2}*{$\rm Method$} &
	\multicolumn{3}{c|}{Text Detection on ICDAR2015/\%} & \multicolumn{3}{c}{Text Tracking on ICDAR2015(video)/\%} \cr\cline{2-7}

	~ & Precision & Recall & F-score & ${\rm ID_{F1}}$/\% & $\rm MOTA$& $\rm MOTP$  \cr\shline \hline
	\hline
	Ours,w/o angle prediction& 73.5 & 65.1 & 68.5  & 36.1 & 16.4 & 68.7  \cr\hline
    Ours,w/ angle prediction(L1 loss)& 84.2 & 82.3 & 83.2  & 54.7 & 43.2 & 74.5  \cr\hline
    Ours,w/ angle prediction(cosine loss)& 86.8 & 81.7 & 84.2(\green{+15.7})  & 57.3(\green{+21.2}) & 44.1(\green{+27.7}) & 75.8(\green{+7.1}) \cr\hline

     \whline
\end{tabular}

	\vspace{-2mm} 
\end{table}

\begin{table}[t]
    \centering
	\caption{\textbf{Ablation study for input query and input image size.} Tracking query bring huge improvement with $3.1\%$ ${\rm ID_{F1}}$.}
	\label{table10}
	\vspace{-2mm} 
	\def\x{{$\footnotesize \times$}}
\scriptsize
\setlength{\tabcolsep}{2pt}
\begin{tabular}{c|ccc|c|cccc}
    \whline
	\multirow{2}*{$\rm Query$} &
	\multicolumn{3}{c|}{Text Tracking on ICDAR2015(video)/\%} & \multirow{2}*{$\rm Short Side$} &
	\multicolumn{4}{c}{Text Tracking on ICDAR2015(video)/\%} \cr\cline{2-4}\cline{6-9}

	~ & ${\rm ID_{F1}}$/\% & $\rm MOTA$& $\rm MOTP$ & ~ & ${\rm ID_{F1}}$/\% & $\rm MOTA$& $\rm MOTP$ & $\rm FPS$  \cr\shline \hline
	\hline
	Obejct query& 54.2 & 40.5  & 76.6 & 480 & 52.6  & 41.5 & 74.5 & 16  \cr\hline
	Obejct + tracking query& 57.3 & 44.1  & 75.8 & 640 & 56.7  & 43.8 & 74.5 & 13  \cr\hline
	- & - & -  & - & 800 & 57.3 & 44.1  & 75.8& 9  \cr\hline

     \whline
\end{tabular}

	\vspace{-2mm} 
\end{table}

\textbf{Angle Prediction.}
The angle prediction and corresponding loss are the main contributions for TransVTSpotter. 
As shown in Table~\ref{table9}, we conduct three experiments to test the effectiveness of the angle prediction and corresponding cosine loss. Without using angle prediction, model with upright-bounding box(two points) results show a dissatisfactory performance(\ie{} $68.5\%$ f-score for detection and $36.1\%$ ${\rm ID_{F1}}$ for tracking). Compared with using angle prediction, there is around $20\%$ performance gap. Besides, using cosine loss present a better performance than the counterpart of L1 loss($84.2\%$ v.s $83.2\%$ for text detection F-score)

\textbf{Tracking Query}
Tracking query is important for our framework, which uses the knowledge of previously detected objects to obtain a set of tracking boxes. 
As shown in Table~\ref{table10}, for only object query, with only learned object query is input as decoder query, and associating the generated detection to track.
This solution achieves a not bad performance by $54.2$ ${\rm ID_{F1}}$. 
But, with tracking query, the performance present further improvement($3.1\%$ ${\rm ID_{F1}}$)

\textbf{Accuracy vs. Speed}
We analyze the inference speed of TransVTSpotter. The time cost is measured using a single Tesla V100 GPU. Table \ref{table10} shows the effect of input image size. With input image size increasing, the model present a better ${\rm ID_{F1}}$ performance from $52.6\%$ to $57.3\%$. When the short-side of the input image is by 800 pixels, the speech and performance all present relatively satisfactory results, so we set it as the default setting in the experiment.


\begin{table}[t]
    \centering
	\caption{\textbf{Experiments for TransVTSpotter on ICDAR2015(video), Minetto and YVT.} `D',`R', `T' and `S' denotes the Detection, Recognition, Tracking, Spotting, respectively. `video' or `image' denote the image or video level dataset.}
	\label{table8}
	\vspace{-2mm} 
	\def\x{{$\footnotesize \times$}}
\scriptsize
\setlength{\tabcolsep}{2pt}
\begin{tabular}{c|ccc|ccc|ccc|ccc}
    \whline
	\multirow{2}*{$\rm Method$} &
	\multicolumn{3}{c}{ICDAR2015(video,T)/\%}&
	
	\multicolumn{3}{c}{ICDAR2015(video,S)/\%}&
	
	\multicolumn{3}{c}{Minetto~\cite{minetto2011snoopertrack}(video,T)/\%}& 
	
	\multicolumn{3}{c}{YVT~\cite{nguyen2014video}(video,T)/\%}
	\cr\cline{2-13}  
	
	
	~ & $\rm MOTA$ & $\rm MOTP$ & ${\rm ID_{F1}}$ & $\rm MOTA$ & $\rm MOTP$ & ${\rm ID_{F1}}$ & $\rm MOTA$ & $\rm MOTP$ & ${\rm ID_{F1}}$ & $\rm MOTA$ & $\rm MOTP$ & ${\rm ID_{F1}}$  \cr\shline \hline
	\hline
	USTB\_TexVideo~\cite{karatzas2015icdar} & 7.4 & 70.7 & 25.9 & 15.6 & 68.5 & 28.2 & - & - & - & - & - & - \cr\hline
	StradVision-1~\cite{karatzas2015icdar}& 7.9&	70.2&	25.8& 8.9 & 70.2 &	31.9 & - & - & - & - & - & -\cr\hline
    USTB\_TexVideo II-2~\cite{karatzas2015icdar}&12.3&	71.8&21.9&13.2&	66.6&	21.3&- & - & - & - & - & - \cr\hline
	AJOU~\cite{karatzas2015icdar,koo2013scene} & 16.4 & 72.7 & 36.1 & - & - & - & - & - & - & - & - & - \cr\hline
	AGD with AGD~\cite{yu2021end}& - & - & - & - & - & - &  75.6 & 74.7 & - & - & - & - \cr\hline
	
	AGD with EAGD~\cite{yu2021end}& - & - & - & - & - & - & 81.3 &  75.7 & - & - & - & - \cr\hline
	Wei \textit{et al.}~\cite{feng2021semantic}& - & - & - & - & - & - &  83.5 & 76.8 & - & - & - & - \cr\hline
	
	Free~\cite{cheng2020free} & 43.2 & \textbf{76.8} & \textbf{57.9} & 52.9 &	74.8 & \textbf{61.8} & - & - & - & \textbf{54.0} & \textbf{78.0} & - \cr\hline
	
	TransVTSpotter(ours)& \textbf{44.1} & 75.8 & 57.3 & \textbf{53.2} & \textbf{74.9} & 61.5 &  \textbf{84.1} & \textbf{77.6} & \textbf{74.7} & 53.9 & 75.9 & \textbf{64.5} \cr\hline

     \whline
\end{tabular}
		
	\vspace{-2mm} 
\end{table}

\subsection{Comparison with State-of-The-Arts for TransVTSpotter}

%
As shown in Table~\ref{table8}, we evaluate TransSpotter on ICDAR2015(video)~\cite{zhou2015icdar}, Minetto(video)~\cite{minetto2011snoopertrack} and YVT(video)~\cite{nguyen2014video} for tracking and video text spotting task. Minetto consists of 5 videos in outdoor scenes. The frame size is 640 x 480 and all videos are used for test when the model training on ICDAR2015(video). Our TransVTSpotter obtains $44.1\%$ and $84.1\%$ for tracking task(MOTA) on ICDAR2015(video) and Minetto, at least $0.6\%$ improvement than the previous models.  

\subsection{Link to Other Video-and-Language Applications}

In this section, we show that the practicability of the proposed BOVText, not a toy benchmark, which can promote other video-and-text application research.

Text spotting in static images has numerous application scenarios: (1) Automatic data entry. SF-Express~\footnote{https://www.sf-express.com/cn/sc/} utilizes OCR techniques to accelerate the data entry process. NEBO~\footnote{https://www.myscript.com/nebo/} performs instant transcription as the user writes down notes. (2) Autonomous vehicle~\cite{mammeri2014road,mammeri2016mser}. Text-embedded panels carry important information, \eg{geo-location, current traffic condition, navigation}, and etc.
(3) Text-based reading comprehension. TextCaps~\cite{sidorov2020textcaps} and text-based VQA~\cite{singh2019towards,biten2019scene} show the new vision-and-language tasks, which need to recognize text, relate it to its visual context, semantic, and visual reasoning between multiple text tokens and visual entities, such as objects. 
Similarly, there are many application demands for video text understanding across various industries and in our daily lives. We list the most outstanding ones that significantly impact, improving our productivity and life quality.
\textbf{Firstly}, automatically describing video with natural language~\cite{xu2016msr,wang2019vatex} can bridge video and language. 
\textbf{Secondly}, video text automatic translation~\footnote{https://translate.google.com/intl/en/about/} can be extremely helpful as people travel, and help video-sharing websites~\footnote{https://www.youtube.com/} to cut down language barriers. 
\textbf{Finally,} text-based video retrieval~\cite{lei2020tvr,liu2020violin} is an irreplaceable business for many companies, such as Google and YouTube.
More details and analyses for application scenarios concerning BOVText in the supplementary material. 

\begin{figure}[t]
\begin{center}
\includegraphics[width=1\textwidth]{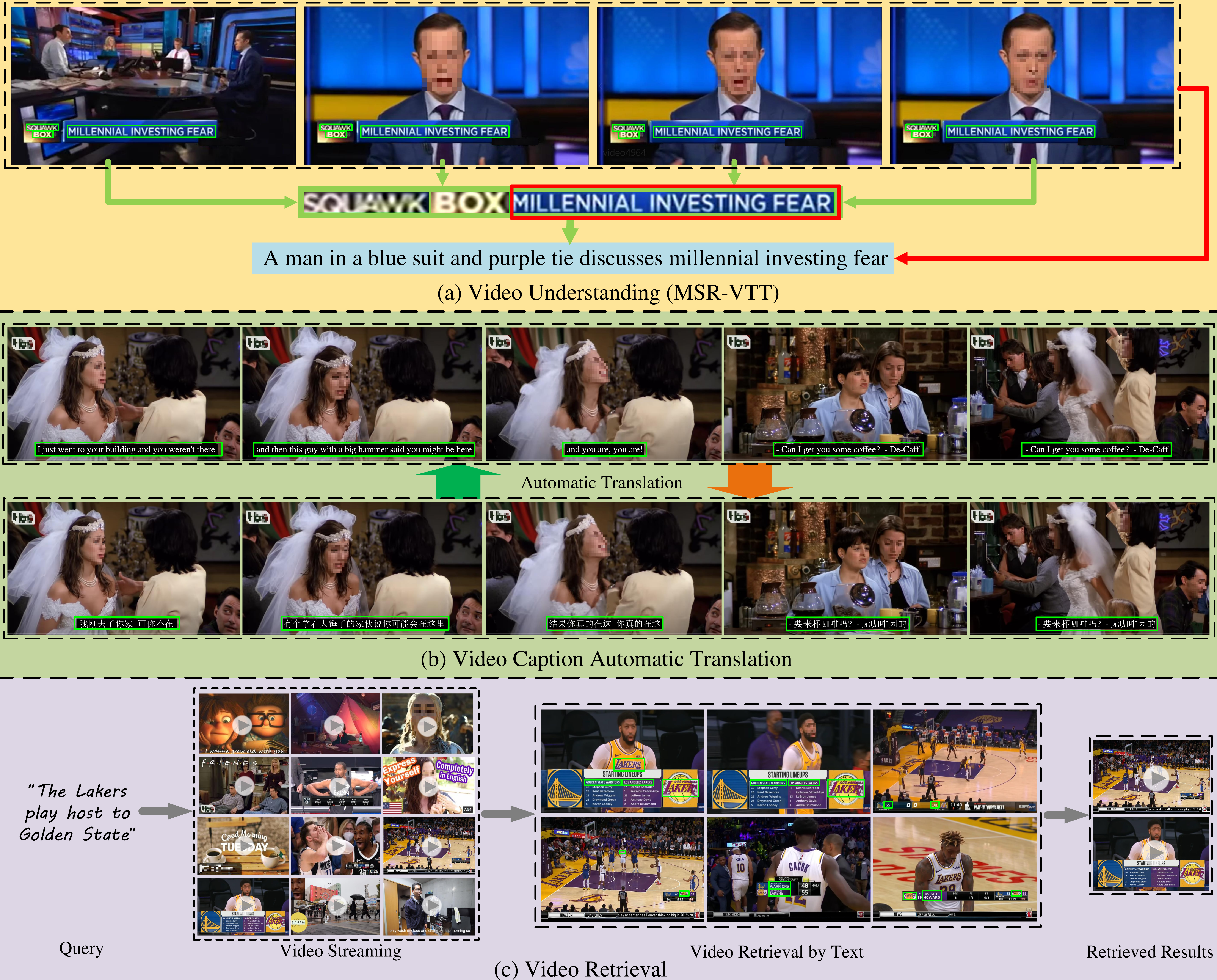}
\vspace{-4mm}
\caption{\textbf{The Real Application Tasks Link to BOVText}. (a) Video Understanding, automatically describing visual content with natural language. (b) Video Caption Translation, extremely helpful for people who travel abroad and video-sharing websites such as YouTube. (c) Video Retrieval, accurate semantic information for text in videos can promote video retrieval.}
\vspace{-4mm}
\label{fig4}
\end{center}
\end{figure}

\textbf{Video Understanding.}
As shown in Figure.~\ref{fig4} (a), the example concerning the task of describing video with natural language is from MSR-VTT~\cite{xu2016msr}, and there has been increasing interest in video understanding~\cite{wang2019vatex,li2016tgif}. However, video description with only visual information is difficult and limited, even for a human. For the annotation of the sample video, \ie{\textit{"A man in a blue suit and purple tie discusses millennial investing fear"}}, we can not learn the information of "millennial investing fear" from the visual information in the video. By comparison, caption and scene texts in the video contain accurate information of \textit{"millennial investing fear"}, which can help the model to describe the video better. We argue that the same as general human understanding, videos without captions and audio, is difficult to be properly understood by the model. We hope the release of BOVText can promote efficient video text reading, further enhancing automatic video description.

\textbf{Video Text Automatic Translation.}
Another practical application is video text automatic translation, as shown in Figure.~\ref{fig4} (b). The application may be unnecessary for several professional teams or classic movies due to the professional translator or huge cost investment. But for international video-sharing websites~\footnote{https://www.youtube.com/}~\footnote{https://www.kuaishou.com/en} with millions of users, it isn't easy to apply multilingual caption and scene text in billions of videos. Therefore, efficient translation concerning caption text (e.g., overlap, song title, logos) and scene text (e.g., street signs, business signs, words on shirt) still need further exploration and research. The large-scale and multilingual BOVText contributes various real scenarios for the development of video text automatic translation.

\textbf{Video Retrieval.}
Video retrieval with textual cues~\cite{mishra2013image,wang2021scene} is also a very important application direction for video-and-text research, as shown in Figure.~\ref{fig4} (c). To the best of my knowledge, video retrieval with text information in the video is still almost a blank field of study and immature application in the industry. The most existing video retrieval methods are stiff combinations of text detection and recognition, invalid for the example with a sentence query. Besides, similar to video understanding, for the query of the sample video, \ie{\textit{"The lakers play host to Golden State"}}, we can not obtain the correct related video without scene text or caption information. The missing information needs to recover by understanding the video with key video text information such as \textit{"GOLDEN STATE WARRIORS, LOS ANGELES LAKERS"}. The proposed BOVText with various text types~(\eg{caption, song title, logos, street signs, business signs}) and annotation can promote the research concerning efficient video retrieval.

\subsubsection{Limitations}
Although the proposed BOVText supports all video text spotting tasks, \ie{\textit{text detection, recognition, tracking end to end video text spotting}}, the potential contributions for other tasks still need mining. For example, as shown in Figure.~\ref{fig4}~(c), we do not provide the corresponding annotation~(\ie{the query for each video}) and metrics concerning video retrieval, but the annotation and metric are easy to obtain due to text spotting annotation already existed. Therefore, there are still many potential contributions for other tasks on BOVText, we want to take these as the future research directions and provide a complete solution method.



\subsection{License and Copyright}
The released video dataset includes two parts: $1,494$ videos from \textit{KuaiShou}~\footnote{https://www.kuaishou.com/en} and $356$ videos from \textit{YouTube}~\footnote{https://www.youtube.com/}. For those videos from \textit{KuaiShou}, we mask the private information such as the human face, which has passed the examination of the legal department and copyright department of KuaiShou corporation. Thus, we own the copyright for these videos. For those videos from \textit{YouTube}, to the best of our knowledge at the time of download, we have exercised caution to download only those videos that were available on YouTube with a Creative Commmons CC-BY (v3.0) License. We don't own the copyright of those videos and provide them for non-commercial research purposes only. All data in our project is open source under CC-by 4.0 license and only be used for research purposes.

\bibliographystyle{plain}
\bibliography{arxiv_v2}

\end{document}